\documentclass[manuscript, anonymous = false, screen]{acmart}

\AtBeginDocument{%
  \providecommand\BibTeX{{%
    \normalfont B\kern-0.5em{\scshape i\kern-0.25em b}\kern-0.8em\TeX}}}

\setcopyright{none}
\copyrightyear{2024}
\acmYear{2024}
\acmDOI{XXXXXXX.XXXXXXX}

\usepackage[utf8]{inputenc}

\begin{document}

\title{Beyond Model Interpretability: Socio-Structural Explanations in Machine Learning}

\author{Andrew Smart}
\affiliation{%
  \institution{Google Research}
  \city{San Francisco}
  \country{USA}}
\email{andrewsmart@google.com}

\author{Atoosa Kasirzadeh}
\affiliation{%
  \institution{Google Research}
  \city{San Francisco}
  \country{USA}}
\email{atoosa.kasirzadeh@gmail.com}

\begin{abstract}  
What is it to interpret the outputs of an opaque machine learning model? One approach is to develop interpretable machine learning techniques. These techniques aim to show how machine learning models function by providing either model-centric local or global explanations, which can be based on mechanistic interpretations (revealing the inner working mechanisms of models) or non-mechanistic approximations (showing input feature-output data relationships). In this paper, we draw on social philosophy to argue that interpreting machine learning outputs in certain normatively-salient domains could require appealing to a third type of explanation that we call ``socio-structural'' explanation. The relevance of this explanation type is motivated by the fact that machine learning models are not isolated entities but are embedded within and shaped by social structures. Socio-structural explanations aim to illustrate how social structures contribute to and partially explain the outputs of machine learning models. We demonstrate the importance of socio-structural explanations by examining a racially biased healthcare allocation algorithm. Our proposal highlights the need for transparency beyond model interpretability: understanding the outputs of machine learning systems could require a broader analysis that extends beyond the understanding of the machine learning model itself.

\end{abstract}

\maketitle

\section{Introduction}

\begin{quote}
    In order to formulate a learning theory of machine learning, it may be necessary to move from seeing an inert model as the machine learner to seeing the human developer—along with, and not separate from, his or her model and surrounding social relations—as the machine learner.
   
    - Reigeluth \& Castelle \cite{reigeluth2021kind}
\end{quote}

The past decade has seen massive research on interpretable machine learning (ML).\footnote{For the purposes of this paper, we use ``interpretable ML,'' ``explainable ML,'' ``interpretable AI,'' and ``explainable AI'' interchangeably.} Here is a rough restatement of the goal of interpretable ML research program: many ML models are \textit{opaque} in that even the expert humans cannot robustly understand, in non-mathematical terms, the reasons for why particular outputs are generated by these models \cite{miller2017explainable,kasirzadeh2021reasons,watson2021local}. To overcome this opacity, various model-centric techniques have been developed to interpret their outputs. These techniques are diverse. They range from producing counterfactual explanations or heatmaps that offer insights into how changing inputs affect outputs \cite{mehdiyev2021explainable,morichetta2019explain,holzinger2022explainable}, to interpreting the inner workings of the model by probing patterns of neuron activations or attention mechanisms \cite{carter2019activation,nanda2023progress,elhage2022toy}.\footnote{Researchers are actively developing unified frameworks that integrate multiple interpretability methods, with the aim of providing a comprehensive conceptual toolkit for understanding the outputs of complex ML models \cite{lundberg2017unified,kasirzadeh2021reasons, schrouff2021best,huber2021local}.}

Despite these advancements, ML interpretability remains a contentious and ambiguous topic in the scientific community, lacking a universally accepted scope and definition \cite{doshi2017towards, lipton2018mythos, molnar2020general, chen2022interpretable}. This ambiguity complicates the evaluation and regulation of opaque ML systems, raising questions about what constitutes sufficient interpretation and how it should be assessed. A pragmatic and pluralistic approach to interpretability has gained traction, viewing explanations as context-dependent responses to why-questions \cite{miller2017explainable,mittelstadt2019explaining,creel2020transparency,kasirzadeh2021reasons}. On this pluralistic approach, the adequacy of an explanation depends on the specific inquiry.

For simple classification tasks, techniques like saliency maps or feature importance may suffice. For instance, if a model is differentiating between images of cats and dogs, saliency maps could highlight the pixels most influential in the decision-making process. However, for complex and socially-embedded topics --- such as biased healthcare algorithms --- these model-centric explanations can fall short. Consider an algorithm that predicts hospital readmission risk but systematically underestimates it for certain racial groups. A model-centric explanation might highlight ``total healthcare costs incurred in the past year'' as an important feature. However, this alone might not fully reveal why the algorithm underestimates risk for a specific racial group. The algorithmic choice could come from the fact that this racial group, due to systemic inequities, have historically been unable to afford adequate healthcare and thus incurred lower costs. As a result, the low value for the ``total healthcare costs incurred in the past year'' feature does not necessarily indicate better health. Instead, it may suggest unmet healthcare needs, leading to higher readmission rates that the algorithm does not effectively account for. In such cases, interpretations that consider both model-specific details like feature importance and relevant social and structural factors like healthcare affordability disparities among racial groups are crucial for understanding ML predictions or decisions.

In this paper, we draw on social philosophy \cite{young2006responsibility,young2010responsibility,epstein2015ant, haslanger2016social, haslanger2020failures} to advocate for a more comprehensive approach to ML interpretability research, expanding beyond model-centric explanations. We propose incorporating relevant socio-structural explanations to achieve a deeper understanding of ML outputs in domains with substantial societal impact. 
In the rest of the paper, we introduce the concept of socio-structural explanations and discuss their relevance to understanding ML outputs. We then examine how these explanations can enhance the interpretation of automated decision-making by ML systems in healthcare \citep{obermeyer2019dissecting}. Our paper expands the discourse on transparency in machine learning, arguing that it extends beyond model interpretability. We propose that in high-stake decision domains, a socio-structural analysis could be necessary to understand system outputs, uncover societal biases, ensure accountability, and guide policy decisions.

\section{Interpretable ML and its discontents}

ML interpretability aims to generate human-understandable explanations for model predictions. This process requires the specification of two key components: the explanandum (the phenomenon requiring explanation) and the explanans (the elements providing the explanation). The model's prediction (or decision) typically serves as the explanandum, while visualizations or linguistic descriptions generated via interpretability techniques act as the explanans. To better understand the landscape of interpretability methods, we provide a broad classification of prominent approaches.\footnote{The interpretable ML literature has grown extensively, making a comprehensive survey beyond the scope of this paper. For recent overviews, see \cite{saeed2023explainable,dwivedi2023explainable,rauker2023toward,bereska2024mechanistic}.}

Model-centric interpretability approaches can be classified according to various criteria, with one fundamental distinction being between intrinsic and post-hoc interpretability \cite{molnar2020interpretable}. Intrinsic interpretability achieves transparency by restricting the complexity of the ML model itself, using approaches such as short decision trees or rule-based systems. In contrast, post-hoc interpretability involves applying methods after model training. These methods include SHAP (SHapley Additive exPlanations) values \cite{lundberg2017unified}, LIME (Local Interpretable Model-agnostic Explanations) \cite{ribeiro2016should}, saliency maps for neural networks \cite{adebayo2018sanity}, and mechanistic interpretability tools \citep{bereska2024mechanistic}.\footnote{Post-hoc methods can also be applied to intrinsically interpretable models, such as computing permutation feature importance for decision trees, which can provide additional insights into their decision-making process.} Another popular classification criterion categorizes ML interpretability techniques into two main types: local and global. This categorization offers a complementary perspective by focusing on the scope and depth of the explanations they provide.

%local
Local explanations focus on explaining individual (or a specific group of) predictions or decisions made by a model. Local explanations often use techniques like feature attribution \cite{ribeiro2016should,lundberg2017unified} or counterfactual instances \citep{mothilal2020explaining,guidotti2022counterfactual}. For example, for an image classification model that predicts "dog," a pixel attribution method might highlight the pixels around the dog's ears and tail as being most influential in the prediction "dog." The explanation could be "The model classified this image as a dog primarily because of the distinctive shapes in these highlighted areas (pointing to highlighted pixels in a visualization). The pointed ears here and the curved tail shape here were the most influential features in making this prediction. Other parts of the image, such as the background or the dog's body, had less impact on the classification." For a loan approval ML model, a counterfactual explanation could be "If your income was 5,000 US dollars higher, your loan would have been approved." 

%global
Global explanations shed light on the average behavior of the model and provide an overall understanding of how a model works across possible inputs. These methods are often expressed as expected values based on the distribution of the data. Global explanations aim to answer questions like "What features are generally most important for this model's predictions?" or "How does the model behave across different types of inputs?" Techniques for global explanations include partial dependence plots \citep{friedman2001greedy} and accumulated local effects \citep{apley2020visualizing}. For example, a partial dependence plot, a type of feature effect plot, can show the expected prediction when all other features are marginalized out. In a house price prediction model, a partial dependence plot might show how the predicted price changes as the house size increases, averaged across all other features like location, number of bedrooms, or age of the house. Since global interpretability methods describe average behavior, they are particularly useful when the modeler wants to understand the general mechanisms in the data, debug a model, or gain insights into its overall performance across various scenarios.

%mechanistic
Mechanistic interpretations expand upon both local and global explanations. These interpretability tools seek to understand the internals of a model. In the case of neural networks, mechanistic interpretability tools reverse engineer the algorithms implemented by neural networks into concepts, often by examining the weights and activations of neural networks. This approach includes methods such as circuit analysis or dictionary learning for identifying specific subnetworks of neurons within larger models to understand the implementation of particular behavior \cite{elhage2021mathematical,templeton2024scaling}.\footnote{Neurons in neural networks can be monosemantic (representing a single concept) or polysemantic (representing multiple unrelated concepts). Monosemantic neurons activate for a single semantic concept, suggesting a one-to-one relationship between neurons and features \citep{templeton2024scaling}. However, neurons are often polysemantic, activating for multiple unrelated concepts, complicating network interpretation. For instance, researchers have empirically shown that for a certain language model, a single neuron can correspond to a mixture of academic citations, English dialogue, HTTP requests, and Korean text \citep{bricken2023monosemanticity}. Polysemanticity makes it difficult to reason about the behavior of the network in terms of the activity of individual neurons.} Mechanistic interpretability is an emerging and highly active area of research, with rapid developments in its neural analysis techniques.

Each of the above-mentioned approaches offers different perspectives on model behavior, ranging from specific instance explanations to overarching principles of operation and fundamental computational mechanisms. The choice of method depends on the specific interpretability goals and the nature of the model being analyzed. There are several acknowledged limitations to existing interpretability approaches. 

First, interpretability techniques can be brittle, sensitive to the target of interpretation \citep{alvarez2018robustness,molnar2020general,watson2022conceptual}, to minor perturbations in model parameters \cite{ghorbani2019interpretation} or input data \cite{rabanser2019failing}. This fragility raises concerns about the reliability and robustness of generated explanations using interpretability methods, especially in real-world scenarios where models are subject to noisy data and evolving conditions. Recent work on mechanistic interpretability has begun to discover features of large language models that are more robust \cite{wang2022interpretability,olsson2022context,templeton2024scaling}. However, there is still significant progress to be made in developing consistently reliable interpretability methods \citep{watson2022conceptual}.

Second, for a given model and input, there may be multiple valid explanations, each highlighting different aspects of the prediction-making or decision-making process \citep{sullivan2024explanation}. This multiplicity embodies both a feature and a bug: as a feature, it reflects the need for a multi-faceted understanding of the complexity of ML predictions; as a bug, it introduces potential confusion and conflicting interpretations, challenging efforts to identify the most relevant or meaningful explanation. Consider, for instance, an ML model used in hiring decisions that recommends not to hire a particular candidate. A feature importance analysis might indicate that the candidate's educational background was the primary factor in the decision. However, a counterfactual explanation might suggest that changing the candidate's gender would alter the outcome. Simultaneously, a SHAP analysis could show that a combination of factors including work experience, interview performance, and age contributed to the decision. Each of these explanations provides insight into the model's reasoning, but emphasizes different aspects, some of which may be more socially sensitive or legally problematic than others. This diversity of explanations challenges practitioners in determining which aspects are most crucial for the model's behavior. Moreover, different stakeholders - such as job applicants, hiring managers, and legal compliance officers - might prefer or trust certain types of explanations over others, further complicating the practical application of these interpretability methods \citep{bhatt2020explainable}. Despite the growing number of interpretability approaches, there is a lack of standardized benchmarks and evaluation frameworks to assess their legal and ethical relevance and compare their performance \cite{agarwal2022openxai, hooker2019benchmark}.

Third, we still have no provable guarantee that a post-hoc explanation accurately reflects the true reasoning behind a model's prediction \cite{rudin2019stop,adebayo2018sanity,watson2022conceptual}. Explanations may be overly simplistic, highlight irrelevant features, or even be misleading, potentially leading to incorrect conclusions about the model's behavior. The potential lack of faithfulness is particularly problematic in high-stakes domains where decisions have significant consequences. For example, a counterfactual explanation for a loan denial might suggest that increasing income would lead to approval. However, the true cause might be a complex interaction of credit history and debt-to-income ratio, not captured by the explanation \citep{kasirzadeh2021use,sullivan2024explanation}. Given these limitations, researchers are exploring novel methods to enhance our understanding of ML models and their predictions (or decisions). 

A promising approach to enhance model transparency involves expanding the scope of interpretations beyond the internal mechanics of the model itself. This expanded perspective recognizes that ML models do not operate in isolation, but within complex social and institutional contexts that can significantly influence their behavior and impact. Here, we propose a new perspective to interpreting ML outputs that incorporates relevant social and structural factors into transparency demands. In particular, we think that in certain situations, the soundness and stability of ML explanations can be improved by appealing to what we call socio-structural explanations that are \emph{external} to an ML model. Our thesis is that in some socially salient applications of ML models, perhaps the most important constraints on model behavior are \emph{external} to the model itself. Extending the idea that the machine learner is not only the inert model, but includes the human developers, uses and surrounding social relations and practices \cite{reigeluth2021kind}, we propose to explain the behavior of a model, in such instances, \textit{given} its place in a \emph{social structure}. We call such explanations \emph{socio-structural} explanations. In order to understand socio-structural explanations, we first need to know what are social structures?

\section{Social structures and socio-structural explanations}

%\subsection{Social structures}  

Social structures are the underlying realities that shape our social lives, influencing our choices, opportunities, and experiences \cite{haslanger2016social,young2006responsibility}. They are the invisible scaffolding of society, both constraining and enabling our individual and collective actions. They give rise to social hierarchies through institutions, policies, economic systems, and cultural or normative belief systems such as race or socioeconomic status \cite{bourgois2017structural}. Social structures manifest in various forms, from the subtle influence of societal norms to the explicit impact of legal frameworks, creating a multilayered reality that shapes our experiences and opportunities.

Social and political philosopher, Iris Marion Young \citep{young2006responsibility,young2010responsibility}, defines social structures as the interplay of institutional rules, interactive routines, resource mobilization, and physical infrastructure. These enduring elements shape the context within which individuals act, offering both opportunities and limitations.\footnote{Social structures, according to Young \cite[p.111]{young2006responsibility}, are defined as follows:
\begin{quote}
As I understand the concept, the confluence of institutional rules and interactive routines, mobilization of resources, as well as physical structures such as buildings and roads. These constitute the historical givens in relation to which individuals act, and which are relatively stable over time. Social structures serve as background conditions for individual actions by presenting actors with options; they provide ``channels'' that both enable action and constrain it.
\end{quote}
} These structures, while socially constructed, possess a reality for exerting tangible influences on individuals and institutions \cite{young2010responsibility, haslanger2012resisting}. They are powerful forces that can constrain and enable actions, cause the specific distribution of resources, and define social roles and expectations. Social structures can explain persistent patterns and circumstances in society, such as racial inequality or gender disparities. To get more concrete, let us analyze the notion of social structures in the context of a socio-structural explanation, borrowing a simple example from Garfinkel \cite{garfinkel1981forms}: 

\begin{quote}
Suppose that, in a class I am teaching, I announce that the course will be ``graded on a curve,'' that is, that I have decided beforehand what the overall distribution of grades is going to be. Let us say, for the sake of the example, that I decide that there will be one A, 24 Bs, and 25 Cs. The finals come in, and let us say Mary gets the A. She wrote an original and thoughtful final.
\end{quote}

\citet{garfinkel1981forms} argues that the explanation "She wrote an original and thoughtful final" is inadequate to answer the explanation-seeking question "Why did Mary get an A?" In a curved grading system, achieving the sole A grade requires more than just quality work. For Mary to earn the only A in the class, her final would need to be the best. If the instructor had not implemented a grading curve, multiple students could have earned As by producing thoughtful and original finals. Garfinkel elaborates on this point, stating "So it is more accurate to answer the question by pointing to the relative fact that Mary wrote the best paper in the class" \cite[p. 41]{garfinkel1981forms}. Mary's A grade was not solely a result of her individual performance, but also a consequence of her relative standing among peers, combined with the specific grading structure that emphasized this comparative aspect. The grading structure, in this case, serves as a crucial contextual element shaping the explanation. More precisely, the structural aspect of this explanation is "the predetermined grading curve that limited the number of As to one," while the social aspect is "Mary's performance relative to her peers (she wrote the best paper in the class)."

Here is a different example for further clarification of the notion of socio-structural explanation. Consider the following explanatory question: ``Why do women continue to be economically disadvantaged relative to men (as opposed to reaching economic parity with men?)'' \cite{haslanger2016social}. \citet{haslanger2016social} argues that we can have (at least) three explanations for this question: biological, individualistic, and structural.

\vspace{5mm}

\textit{Biologistic explanation}: Women are inherently less capable than men in biological qualities deemed necessary (such as intelligence or competitiveness) for success in high-paying jobs.

\vspace{5mm}

\textit{Individualistic explanation}: Women, to a greater extent than men, prioritize child-rearing over pursuing high-paying careers, thus voluntarily sacrificing economic success for the perceived rewards and satisfactions of motherhood.

\vspace{5mm}

\textit{Structural explanation}: Women are embedded within a self-reinforcing economic structure that systematically disadvantages them through institutional practices, social norms, and power dynamics.

\vspace{5mm}

Each of these explanations refers to different causes, operating at distinct levels of analysis. The biologistic and individualistic explanations focus on factors intrinsic to individuals or groups, without considering the broader socio-structural context. In contrast, the structural explanation situates individual actions and outcomes within a larger system of interconnected social forces. If the social structure is in place, then we can view individuals as occupying specific "nodes" within a complex social network or structure. The socio-structural explanation posits that gender wage disparities arise from the complex interplay of societal, economic, and institutional factors that collectively shape opportunities and constraints. Given the socio-structural limitations in place, we can explain why women, at the population level, experience economic disadvantages compared to men based on their position within the social structure. In this context, "social structure" refers to the complex network of institutions, relationships, and cultural norms that organize society. It includes economic systems that historically undervalue work traditionally performed by women, political institutions that may underrepresent women's interests, and educational structures that can reinforce gender stereotypes. Additionally, it includes cultural norms that influence career choices and work-life balance expectations, organizational hierarchies that often favor male leadership, and legal frameworks that may inadequately address gender discrimination. 

Women's place within this multifaceted social structure often results in reduced access to resources, limited decision-making power, and fewer opportunities for advancement, collectively contributing to persistent economic disparities at the population level. The socio-structural approach to explanation, when rigorously applied, offers valuable insights. It demonstrates how individual choices and actions can be profoundly shaped by the surrounding social structures. By highlighting the influence of broader structural forces on seemingly personal decisions, it reveals patterns often operating beyond an individual's immediate awareness or control. 

Let us draw a close analogy between the above instances of socio-structural explanations and a toy example of interpreting an ML model's output. Consider an ML-powered hiring model that consistently recommends male candidates over female candidates for senior executive positions in a tech company. An initial explanation of the recommendations generated by a SHAP interpretability method might say: "The model recommends male X over female Y because X's features contribute more positively to the model's output. Specifically, X's 10 years of tech leadership experience contributes +0.4 to the score, while Y's 7 years contributes only +0.2." Let us assume similar explanations (relating years of tech leadership experience to the recommendation score) are generated for a population of females. These explanations might fail to capture the full picture. 

%

%For instance, the auditors find that the model fails to account for historical gender biases in tech leadership roles, differences in networking opportunities, and variations in how leadership qualities are perceived across genders. Additionally, factors such as geographic disparities affecting job availability and commute times --- which disproportionately impact women due to societal expectations around family care --- are entirely absent from the model's considerations. 

Upon deeper investigation, an auditor team uncovers a more complex and nuanced reality. First, the auditors find that the ML model was trained on the company's historical hiring data from 2000-2020, which included 85\% male executives. This data reflects the company's past hiring practices, which favored men for leadership roles. The socio-structural aspect here is the historical underrepresentation of women in executive positions, rooted in long-standing societal norms and institutional practices. A socio-structural explanation could look like: "The model's bias reflects decades of systemic exclusion of women from leadership roles in the tech industry, perpetuating a cycle where the lack of female representation in executive positions reinforces the perception that these roles are best suited for men." Second, the ML model places high importance on continuous work experience, with any gap longer than 6 months reducing a candidate's score by 0.1 per year. 40\% of female candidates had career gaps averaging 2.5 years, compared to 10\% of male candidates averaging 1 year, often coinciding with childbearing ages. This reflects the socio-structural reality of women bearing a disproportionate responsibility for child-rearing and family care, leading to more frequent and longer career interruptions. A socio-structural explanation could look like: "The model's penalty for career gaps disproportionately impacts women due to societal expectations and norms that place the primary burden of childcare and family responsibilities on women, resulting in more frequent and longer career interruptions that are then interpreted by the model as reduced qualifications." Third, the model does not consider geographic location in its evaluation. However, geographic disparities affect job availability and commute times, disproportionately impacting women with childbearing responsibilities. This reflects socio-structural expectations around family care that often limit women's job options to those closer to home or with flexible hours. A socio-structural explanation could look like: "The model's failure to account for geographic factors overlooks the societal expectations that often constrain women's job choices based on proximity to home and flexibility for family care. This oversight particularly disadvantages women who may be highly qualified but limited in their job options due to these socially imposed constraints."

Producing rigorous socio-structural explanations can be challenging as it requires significant sociological understanding and interdisciplinary expertise. However, once obtained, these explanations enable novel forms of interventions. Here are some examples of possible interventions enabled by obtaining socio-structural explanations. The first is to modify the model to cap the maximum contribution of "continuous experience" at +0.2. The second is to introduce a new feature "diverse experience" that values varied career paths, including those with gaps. The third is to augment the training data with 500 profiles of successful executives who have had career gaps of 1-3 years, ensuring at least 50\% are women. The fourth is to implement a company-wide policy requiring human review for any candidate the ML system ranks lower primarily due to career gaps (>0.2 score difference). This toy example is supposed to highlight that integrating socio-structural explanations into the ML transparency toolkit enables us to transcend superficial model-centric solutions (when relevant) and address the fundamental causes underlying ML outputs.

Of course, the specific interventions depend on what we want to change and the particular context of the problem at hand. Socio-structural explanations are not always useful or applicable in every situation. The effectiveness of these explanations and subsequent interventions can vary based on the complexity of the social systems involved, the quality of available data, and the specific goals of the analysis. In some cases, other approaches might be more appropriate or effective. 

In the following section, we examine a case study of algorithmic deployment in healthcare decision-making, highlighting the critical relevance of socio-structural explanations in this context. This analysis demonstrates how a deeper understanding of social structures can inform more effective strategies for developing and implementing algorithmic systems in high-stakes decision domains. While our original focus was on socio-structural explanations for ML systems, we recognize that the importance of these explanations generalizes to a broader range of automated decision systems.

\section{Socio-structural explanations of racial bias in health-care algorithms}

A widely discussed example in the growing body of literature on algorithmic bias is the study by \citet{obermeyer2019dissecting}. This research revealed that a commonly used US hospital predictive algorithm for allocating scarce healthcare resources systematically discriminated against Black patients. Specifically, the algorithm assigned lower risk scores to Black patients who were equally in need as their White counterparts. The root cause was the algorithm's use of healthcare costs as a proxy for "healthcare need" \cite{ledford2019millions}. This approach led to a significant underestimation of health risks for Black patients who, on average, incurred lower healthcare costs than White patients with similar chronic conditions due to systemic disparities in care access and quality \cite{obermeyer2019dissecting}.

Empirical investigations demonstrated that the care provided to Black patients cost an average of USD 1,800 less per year than the care given to a white person with the same number of chronic health problems. At a given risk score, Black patients are considerably sicker than White patients, as evidenced by signs of uncontrolled illnesses \cite{obermeyer2019dissecting}. The algorithm predicted that this disparity in spending corresponded to a similar disparity in actual health-care needs and therefore risk score. Consequently, Black people had to be much sicker in order to be referred for treatment or other resources. The algorithm's prediction of health needs is, in fact, a prediction on health costs \cite{obermeyer2019dissecting}.

When algorithmic decision systems fail in consequential domains like health-care, the repercussions can be severe, potentially leading to patient deaths. It is crucial to understand the reasons and causes for such failures. Therefore, explaining the "why" behind these failures through the analysis of failed outputs is critical. One prevalent type of failure is algorithmic bias that perpetuates existing socio-structural inequalities, such as structural racism \cite{hanna2020towards, obermeyer2019dissecting, raji2020closing, barocas-hardt-narayanan}. Structural racism refers to the complex ways in which historical and contemporary racial inequities are reproduced through interconnected societal systems like healthcare, education, housing, and the criminal justice system \cite{pallok2019structural}. Even when race is not explicitly considered, its influence can be deeply embedded in the data, shaping associations and outcomes \cite{robinson2020teaching}. In the context of this instance, the following explanatory question demands a response: Why did this algorithm systematically discriminate against Black people?

To answer this question, we must consider both the interpretation in reference to the model and the broader socio-structural context in which it operates.\footnote{For a discussion of the levels of interpretation see \citet{creel2020transparency} and \citet{kasirzadeh2021ethical}.} \citet{obermeyer2019dissecting} show that this particular algorithm discriminated against Black patients due to its use of healthcare costs as a proxy for "healthcare need." This choice reflects a fundamental misunderstanding of the relationship between costs and needs in a healthcare system marked by systemic racial disparities. \citet{obermeyer2019dissecting} demonstrated that conditioning on healthcare costs is the mechanism by which the bias arises in this case, and we must change the data we feed the algorithm and use new labels that better reflect social reality, which in turn requires deep understanding of the domain, the ability to identify and extract relevant data elements, and the capacity to iterate and experiment \cite{obermeyer2019dissecting}. The socio-structural interpretation of the algorithm's behavior is as follows. 

\begin{quote}
The algorithm discriminates against Black patients because it is designed and deployed in a healthcare system characterized by longstanding racial inequities. By encoding healthcare costs as a proxy for healthcare needs, the algorithm inadvertently encodes and perpetuates systemic disparities in care access and quality. Black patients, on average, incur lower healthcare costs not because they are healthier, but due to historical patterns of exclusion, lack of access to care, and underinvestment in healthcare resources for Black communities. The algorithm interprets these lower costs as lower needs, thereby underestimating the health risks for Black patients and perpetuating a cycle of inadequate care allocation. This reflects how the algorithm manages to reproduce structural racism through its uncritical use of data that embodies these systemic inequalities. 
\end{quote}

It remains challenging for practitioners to identify the harmful repercussions of their own systems prior to deployment, and, once deployed, emergent issues can become difficult or impossible to trace back to their source \cite{raji2020closing}. Unfortunately, many failures of algorithmic decision systems in the healthcare industry disproportionately impact people or communities who have been put already in a structurally vulnerable social positions \cite{leslie2021does}. This can be due to many factors. However, a consistent theme in the study of these failures, that is often only revealed after the fact, is that there is a lack of socio-structural understanding among the designers and users of these systems \cite{martin2020participatory}. The study presented in this section exemplifies this challenge. Employing model-centric explanations would likely highlight the importance of the cost feature to algorithmic output, but would not expose the underlying racial bias originating from historical and systemic inequalities in healthcare access and delivery. In this context, socio-structural explanations consider the relevant societal context in which the model operates, in relation to relevant historical biases, societal norms, and institutional practices.

\section{Implications for ML transparency research and conclusion}

%We have argued that machine learning transparency research implicitly adopts model-centric approaches when attempting to explain or interpret models. This means an assumption that the best interpretation of the model's behavior can be found \textit{inside} the model in terms of \textit{local} or \textit{global} explanations. We have introduced a third kind of explanation: \textit{socio-structural} explanations to ML transparency research. These kinds of explanations view the model as occupying certain positions within larger social structures, and that these structures - rather than internal states of the model - constrain or enable the model's behavior. This implies that the ML model cannot be seen as separate from its developers and the surrounding social relations and practices. 

ML research and practice are fundamentally shaped by the approaches adopted by practitioners. These approaches influence the entire process: from the questions asked and data collected, to the choice of objective functions and the selection of proxy or target variables for optimization. Throughout this paper, we have argued that model-centric explanations, while valuable, can be inadequate for comprehensively understanding whether a model truly benefits or potentially harms people. This inadequacy is particularly pronounced in high-stakes domains where ML models are often developed and deployed into complex social and structural contexts without sufficient domain-specific theoretical understanding. We have argued that to meaningfully interpret the \textit{social} predictions (or decisions) of models in high-stake domains, a deep socio-structural understanding is required.

One challenge lies in that many ML practitioners and researchers may not feel adequately equipped to analyze and respond to social structures. Alternatively, they may be hindered from leveraging social structural knowledge due to constraints in time, training, incentives, or resources \cite{shankar2022operationalizing}. This gap between technical expertise and socio-structural understanding presents a significant hurdle in developing truly beneficial ML systems.
Algorithmic transparency and accountability research in ML is often motivated by the need to foster trust in these systems. Much of this research rightly argues for the critical importance of model-centric interpretations \cite{ghorbani2019interpretation, han2022explanation, lundberg2017unified}. However, the demands for transparency of ML models must extend beyond model-centric details to encompass socio-structural factors in socially-salient prediction or decision domains.
Producing new, more representative labels and objectives for ML models requires a deep understanding of the domain, the ability to identify and extract relevant data elements, and the capacity to iterate and experiment \cite{obermeyer2019dissecting}.

The importance of socio-structures has been increasingly recognized in recent literature on algorithmic justice \citep{hoffmann2019fairness,kasirzadeh2022algorithmic,lin2022artificial}. These works argue for a more holistic approach to ML development and deployment, one that considers not just model-centric measures but also societal impacts.
In light of these considerations, we call for further research into the integration of socio-structural understanding into different stages of the ML lifecycle, from problem formulation and data collection to model development, deployment, and ongoing monitoring. We think that sometimes socio-structural interpretations can reveal causally-relevant reasons for why an algorithm behave in a certain way for a certain population. By doing so, we can work towards ML systems that are not only technically proficient but also socially aware and beneficially aligned.

\section{Acknowledgments}
We would like to thank Donald Martin, Been Kim, Darlene Neal, Mayo Clinic Accelerate Program and The Impact Lab team at Google Research for helpful discussions and feedback on earlier drafts of this paper.

\bibliographystyle{ACM-Reference-Format}
\bibliography{think_outside_the_black_box.bib}

\end{document}